\documentclass[10pt,conference]{IEEEtran}
\IEEEoverridecommandlockouts
\usepackage{cite}
\usepackage{amsmath,amssymb,amsfonts}

\usepackage{algorithmic}
\usepackage{algorithm} 

\usepackage{graphicx}
\usepackage{textcomp}
\usepackage{balance}   
\usepackage{url}       

\usepackage{booktabs}  
\usepackage{multirow}  
\usepackage{tabularx}  

\usepackage[table]{xcolor} 
\usepackage{arydshln}

\usepackage[hidelinks]{hyperref} 

\usepackage{tcolorbox}
\tcbuselibrary{skins} 
\usepackage{xcolor}
\usepackage{enumitem}

\definecolor{mBlue}{RGB}{105, 129, 168}  
\definecolor{mBlueBack}{RGB}{240, 244, 250} 

\definecolor{mRed}{RGB}{186, 126, 126}
\definecolor{mRedBack}{RGB}{252, 245, 245}

\newtcolorbox{promptbox}[2][]{
  enhanced,
  title={#2},
  colframe=#1,          
  colbacktitle=#1,      
  colback=#1Back,       
  coltitle=white,       
  fonttitle=\bfseries\small,
  fontupper=\footnotesize\sffamily, 
  boxrule=0.5mm,        
  arc=1.5mm,            
  attach boxed title to top left={yshift=-2mm, xshift=3mm}, 
  boxed title style={boxrule=0pt, arc=1mm, frame code={}},
  top=3mm, bottom=2mm, left=2mm, right=2mm,
  before skip=4pt, after skip=4pt 
}

\newtcolorbox{outerbox}[1][]{
  enhanced,
  colback=white,        
  colframe=gray!60,     
  title={#1},
  coltitle=black,
  fonttitle=\bfseries\small,
  attach boxed title to top center={yshift=-2mm},
  boxed title style={colback=white, boxrule=0pt, frame code={}}, 
  boxrule=0.3mm,
  arc=2mm,
  top=4mm, bottom=2mm, left=2mm, right=2mm,
}

\definecolor{morandiBlue}{RGB}{176, 196, 222} 
\definecolor{morandiGrayGreen}{RGB}{188, 210, 188} 
\definecolor{morandiPink}{RGB}{222, 188, 188} 
\definecolor{morandiBeige}{RGB}{210, 205, 185} 
\definecolor{MorandiOrange}{RGB}{240,210,190}

\def\BibTeX{{\rm B\kern-.05em{\sc i\kern-.025em b}\kern-.08em
    T\kern-.1667em\lower.7ex\hbox{E}\kern-.125emX}}
\begin{document}

\title{ABEX-RAT: Synergizing Abstractive Augmentation and Adversarial Training for Classification of Occupational Accident Reports
}


\author{
    \IEEEauthorblockN{
        Jian Chen\IEEEauthorrefmark{1}\thanks{*Corresponding author.},
        Jiabao Dou\IEEEauthorrefmark{2}
    }
    \IEEEauthorblockA{\IEEEauthorrefmark{1}Ningxia Research Institute of Transport Science, Yinchuan, China}
    \IEEEauthorblockA{\IEEEauthorrefmark{2}Department of Computer Science, Hong Kong Baptist University, Hong Kong}
}

\maketitle

\begin{abstract}
The automatic classification of occupational accident reports is pivotal for workplace safety analysis but is persistently hindered by severe class imbalance and data scarcity. In this paper, we propose \texttt{ABEX-RAT}, a resource-efficient framework that synergizes generative data augmentation with robust adversarial learning. Unlike computationally expensive large language models (LLMs) fine-tuning, our approach employs a two-stage abstractive-expansive (ABEX) pipeline: it first utilizes a prompt-guided LLM to distill label-critical semantics into concise abstracts, which are then expanded into diverse synthetic samples to balance the data distribution. Subsequently, we train a lightweight classifier using a random adversarial training (RAT) protocol, which stochastically injects perturbations to enhance generalization without significant computational overhead. Experimental results on the OSHA dataset demonstrate that ABEX-RAT establishes a new state-of-the-art, achieving a Macro-F1 score of 90.32\% and significantly outperforming both traditional baselines and fine-tuned large models. This confirms that targeted augmentation combined with robust training offers a superior, data-efficient alternative for specialized domain classification. The source code will be made publicly
available upon acceptance.
\end{abstract}

\begin{IEEEkeywords}
Occupational Safety, Text Classification, Data Augmentation, Large Language Models, Adversarial Training.
\end{IEEEkeywords}

\begin{figure}[h]
\centering
\includegraphics[width=0.45\textwidth]{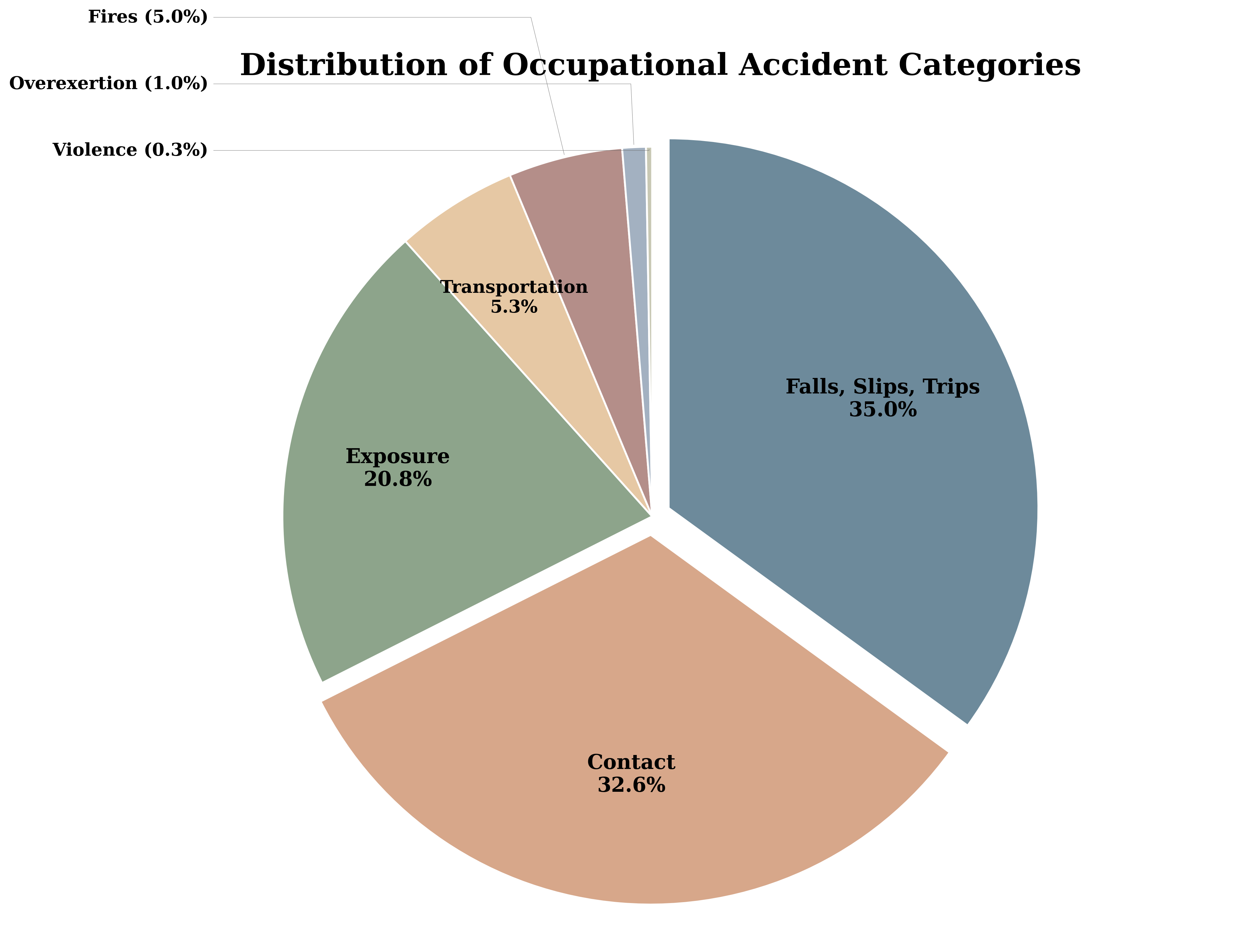}
\caption{The problem of category imbalance in occupational accident reports: a case study of OSHA dataset.}
\end{figure}

\begin{figure*}
\centering
\includegraphics[width=0.9\textwidth]{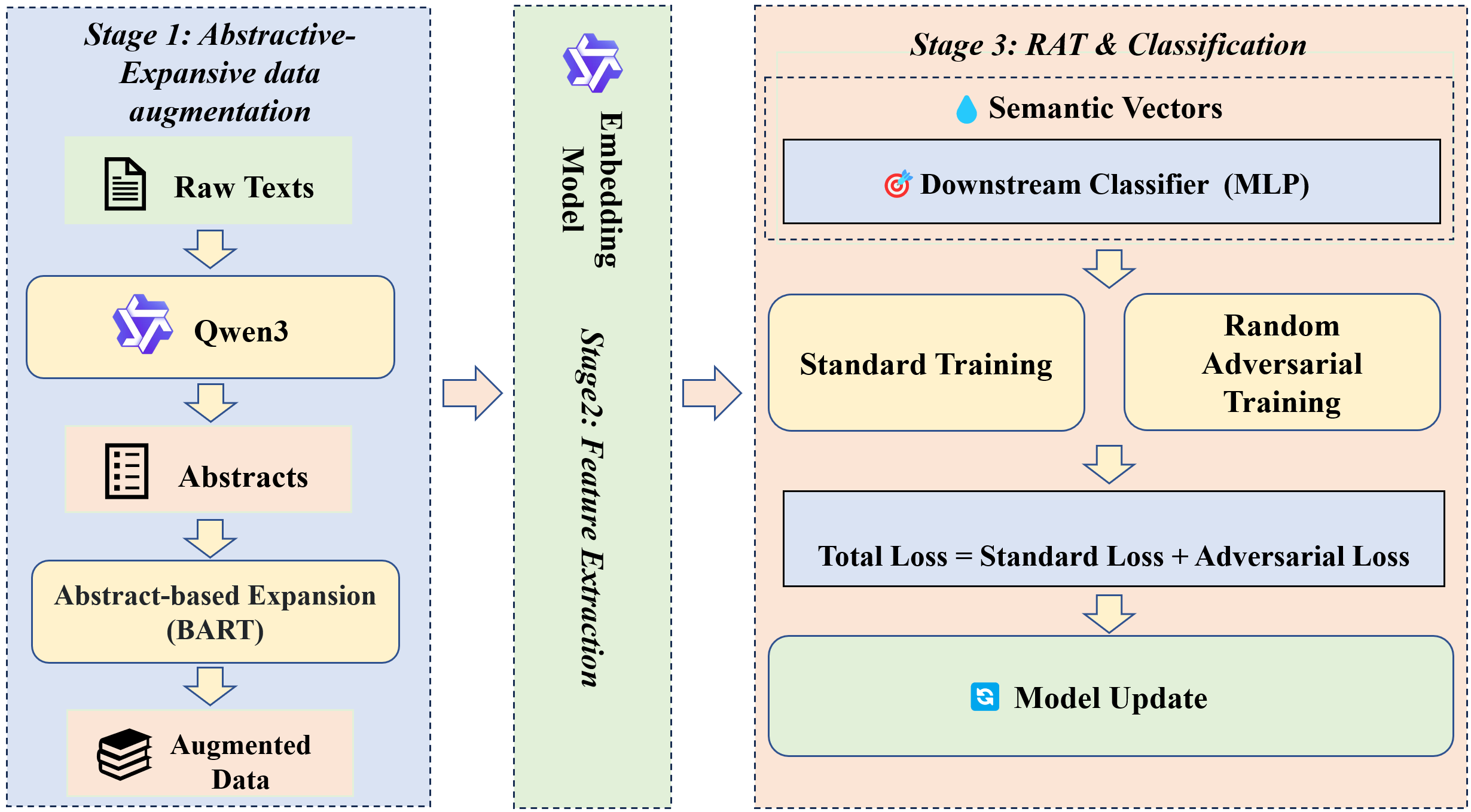}
\caption{The overall architecture of our proposed ABEX-RAT framework. The framework consists of three stages. \textbf{Stage 1 (ABEX Data Augmentation):} A large language model (e.g., Qwen3) generates a concise abstract from each raw text, which is then expanded into multiple augmented samples by a BART-based model. \textbf{Stage 2 (Feature Extraction):} A pre-trained embedding model converts all texts into dense semantic vectors. \textbf{Stage 3 (RAT \& Classification):} A lightweight MLP classifier is trained on these vectors using a total loss that stochastically combines a standard loss with an adversarial loss to improve model robustness.}
\end{figure*}

\section{Introduction}

The analysis of occupational accident reports is a cornerstone of workplace safety management, providing critical insights that help prevent future incidents \cite{tixier2016automated,zhang2019construction,qiao2022construction}. Automatic classification of these free-text narratives into predefined categories is a crucial task, as it enables large-scale statistical analysis and the identification of risk patterns. However, this task is fraught with challenges, chief among them being the severe class imbalance inherent in real-world data \cite{shuang2024INSTRUCTOR-CIT}, as illustrated in Figure 1. Catastrophic but rare events (e.g., \textit{Fires and Explosions}) are vastly outnumbered by more common accident types, leading to models that are biased towards the majority classes and exhibit poor predictive performance on the very incidents that may require the most urgent attention \cite{shayboun2025review}. While recent works have explored LLM-based agents for safety monitoring \cite{li2025safetygpt}, stable classification of rare accident types remains an open challenge due to the lack of specialized training data.

Prior research has predominantly focused on two paradigms to address this challenge: model-centric adaptation and data-centric augmentation. In the model-centric domain, researchers have attempted to fine-tune pre-trained language models (PLMs) such as BERT and RoBERTa \cite{devlin2019bert, liu2019roberta}. While these models capture rich semantic features, they often struggle to generalize when faced with extreme data scarcity for minority classes \cite{khallaf2021classification}. To mitigate this, cost-sensitive learning strategies, such as the focal loss \cite{lin2017focal} or re-sampling techniques like SMOTE \cite{chawla2002smote}, have been employed to assign higher weights to hard-to-classify examples. However, re-weighting schemes alone cannot compensate for the fundamental lack of feature diversity in underrepresented classes \cite{chen2025multimodal}.

On the data-centric front, traditional data augmentation techniques have been widely used \cite{wei2019eda,sennrich2016improving}. Yet, these methods often produce low-quality samples that lack semantic coherence. More recently, the advent of large language models (LLMs) \cite{achiam2023gpt4,ouyang2022training} has opened new avenues for generative data augmentation. LLMs demonstrate remarkable capability in generating fluent and diverse text. However, our empirical findings indicate that zero-shot performance of LLMs is often suboptimal for specialized domains like occupational safety due to the lack of domain-specific knowledge \cite{shumailov2023curse}. Furthermore, brute-force fine-tuning of these massive models for data generation is computationally prohibitive for many industrial applications \cite{chen2026qwen}.

Parallel to data augmentation, adversarial training (AT) has emerged as a powerful regularization technique to enhance model robustness and generalization \cite{goodfellow2014explaining}. Methods like the fast gradient method (FGM) \cite{fgm} and projected gradient descent (PGD) \cite{madry2017towards} operate by introducing small perturbations to the input embeddings, forcing the model to learn smoother decision boundaries. While effective, standard AT methods typically require multiple backward passes to generate perturbations, significantly increasing the training time cost \cite{chen2025rat}. Consequently, integrating AT into large-scale text classification tasks remains a challenge, particularly when efficiency is a priority.

To address these interconnected limitations—data scarcity, semantic diversity, and training efficiency—we propose \texttt{ABEX-RAT}, a novel framework that synergizes a sophisticated data augmentation pipeline with a robust training strategy. Instead of relying on costly full-parameter fine-tuning of LLMs, our approach first enriches the dataset using an abstractive-expansive (ABEX) data augmentation process, inspired by recent advances in generative models \cite{ghosh2024abex}. By leveraging a prompt-guided LLM to distill core semantics and a generative model to expand them \cite{lewis2020bart}, ABEX generates diverse, high-quality synthetic samples specifically for minority classes. 

Subsequently, to handle the augmented feature space robustly, we train a lightweight classifier using a computationally efficient random adversarial training (RAT) protocol \cite{chen2025rat}. Unlike standard AT, RAT applies perturbations stochastically, thereby enhancing model robustness and generalization without imposing significant computational overhead. This synergistic approach effectively bridges the gap between the need for diverse training data and efficient, robust model optimization.

The main contributions of this work are summarized as follows:
\begin{itemize}
    \item We propose \texttt{ABEX-RAT}, a novel framework that synergizes two components: the ABEX pipeline for generative data augmentation to mitigate class imbalance, and the RAT protocol for computationally efficient adversarial training to enhance model robustness.
    \item We introduce a specialized prompt-based abstraction and expansion mechanism that ensures the semantic consistency of synthetic accident reports, addressing the hallucinations common in direct LLM generation.
    \item We demonstrate that on the public OSHA dataset, our method establishes a new state-of-the-art (SOTA), reaching a Macro-F1 score of 90.32\% and significantly outperforming traditional fine-tuned models, zero-shot LLMs, and previous SOTA approaches.
\end{itemize}

\section{METHODOLOGY}

To address the challenges of classifying occupational accident reports—specifically the dual issues of data scarcity in critical categories and severe class imbalance—we propose the ABEX-RAT framework. This framework synergizes a generative data augmentation pipeline with a computation-efficient robust training strategy. The overall architecture, illustrated in Figure 2, operates as a three-stage pipeline: (1) abstractive-expansive data augmentation to balance the dataset; (2) embedding-based feature extraction; and (3) random adversarial training for robust classification.

\subsection{Stage 1: ABEX Data Augmentation}
The core hypothesis of ABEX is that direct augmentation of raw text often introduces noise or fails to capture the causal logic of accidents. Therefore, we decompose the augmentation process into two sequential steps: \textit{abstraction} and \textit{expansion}.

\subsubsection{Prompt-Guided Abstraction}
First, each raw accident report, $T_{\text{raw}}$, is distilled into a concise abstract, $T_{\text{abs}}$, using a LLM. We utilize Qwen3-Instruct\footnote{\url{https://www.modelscope.cn/models/Qwen/Qwen3-235B-A22B-Instruct-2507}} as the backbone. Unlike generic summarization, our abstraction function $\mathcal{A}$ is guided by a domain-specific prompt $\mathcal{P}$ designed to retain only the label-defining semantics (e.g., hazard source, incident mechanism, and injury type) while discarding irrelevant stylistic variations and potential personally identifiable information.

\begin{equation}
    T_{\text{abs}} = \mathcal{A}_{\text{LLM}, \mathcal{P}}(T_{\text{raw}})
\end{equation}

To ensure high-quality abstraction, we design a structured prompt $\mathcal{P}$ that acts as a strict instruction set for the LLM. As visualized in Figure 3, the prompt enforces constraints on the output style and explicitly requires the preservation of key concepts specific to the accident category. By constraining the LLM to output a structured summary, we standardize the semantic representation across heterogeneous report styles.

\begin{figure}
    \centering
    \begin{promptbox}[mBlue]{Prompt Template for Abstraction Stage}
        \textbf{\# Role}\\
        You are an expert in data augmentation, specializing in creating abstract descriptions of incident reports for machine learning purposes.
        
        \vspace{1mm}
        \textbf{\# Task}\\
        Your task is to convert the provided "Original Incident Report" into a short, abstract "Abstract Description". This description must:
        \begin{enumerate}[leftmargin=*, nolistsep]
            \item Retain the core event and intent of the report.
            \item Remove specific, non-essential details like personal names, exact locations, specific dates, and company names unless they are critical to the core event.
            \item Crucially, you MUST preserve the core meaning of the following "Key Concepts" which are vital for this incident's category classification.
        \end{enumerate}

        \vspace{1mm}
        \textbf{\# Constraints}\\
        \textbf{- Category:} \{category\}\\
        \textbf{- Key Concepts to Preserve:} \{keywords\}\\
        \textbf{- Output Style:} A concise, general, and factual summary of the core event.

        \vspace{1mm}
        \textbf{\# Task Execution}\\
        \textbf{- Original Incident Report:}\\
        \texttt{"""\\
        \{full\_text\}\\
        """}\\
        \textbf{- Abstract Description:}
    \end{promptbox}
    \caption{The specific prompt design used in the ABEX abstraction phase. The placeholders \{category\}, \{keywords\}, and \{full\_text\} are dynamically filled during inference.}
    \label{fig:prompt_design}
\end{figure}

\subsubsection{Diversity-Driven Expansion}
Next, to generate synthetic samples, we employ a pre-trained generative model\footnote{\url{https://huggingface.co/utkarsh4430/ABEX-abstract-expand}}~\cite{ghosh2024abex} as the expansion function $\mathcal{E}$. This model takes the concise abstract $T_{\text{abs}}$ and hallucinates plausible, grammatically diverse full-text variations, effectively reversing the abstraction process. 

To explicitly address class imbalance, the number of generated samples is dynamically calculated. Let $N_c$ be the number of original samples in class $c$, and $N_{\text{max}}$ be the count of the majority class. The expansion factor $R_c$ for class $c$ is defined as:
\begin{equation}
    R_c = \left\lceil \lambda \cdot \frac{N_{\text{max}}}{N_c} \right\rceil
\end{equation}
where $\lambda \in (0, 1]$ is a hyperparameter controlling the degree of balancing. This ensures that minority classes receive significantly more synthetic samples:
\begin{equation}
    \{T_{\text{aug}}^{(1)}, \dots, T_{\text{aug}}^{(R_c)}\} = \mathcal{E}_{\text{BART}}(T_{\text{abs}})
\end{equation}

\subsection{Stage 2: Feature Extraction}
Following augmentation, we adopt an efficient embedding-based paradigm rather than fine-tuning the entire LLM. We utilize Qwen3-Embedding model~\cite{qwen3embedding} as a fixed feature extractor $\Phi$. This model is optimized for semantic retrieval and clustering, making it ideal for separating confusing accident categories. Each text sample $T_i$ is mapped to a dense vector $\mathbf{x}_i$:
\begin{equation}
    \mathbf{x}_i = \Phi(T_i) \in \mathbb{R}^d
\end{equation}
where $d=4096$. This process yields a dataset of embedding-label pairs, $\mathcal{D}_{\text{emb}} = \{(\mathbf{x}_i, y_i)\}$, which serves as the static input for the classifier. This design freezes the massive parameters of the LLM, significantly reducing the memory footprint during the training of the downstream classifier.

\begin{table*}
\centering
\caption{Results comparison with different methods. \textbf{Bolded} values indicate the best performance.}

\setlength{\aboverulesep}{0pt}
\setlength{\belowrulesep}{0pt}
\renewcommand{\arraystretch}{1.25}

\begin{tabularx}{0.9\textwidth}{@{}Xcccccc@{}}
\toprule

\rowcolor{gray!25}
\textbf{Models} & \multicolumn{3}{c}{\textbf{Weighted-F1 (\%)}} &\multicolumn{3}{c@{}}{\textbf{Macro-F1 (\%)}} \\ 

\cmidrule(r){2-4} \cmidrule(l){5-7} 

\rowcolor{gray!25}
 & Precision   & Recall  & \textbf{Weighted-F1}    & Precision   & Recall   & \textbf{Macro-F1} \\

\midrule
\multicolumn{7}{c}{\textit{Traditional Methods}} \\
FastText \cite{joulin2016fasttext} & 88.62 & 88.59 & 88.59 & 68.29 & 68.52 & 68.39 \\
BERT-base-uncased \cite{devlin2019bert} & 91.48 & 91.50 & 91.47 & 88.97 & 87.93 & 88.43 \\
RoBERTa-base \cite{liu2019roberta} & 90.90 & 90.83 & 90.82 & 91.92 & 88.11 & 89.74 \\
\hdashline
Llama3.1-8B-Instruct \cite{grattafiori2024llama3} & 72.67 & 68.90 & 68.93 & 49.83 & 49.63 & 46.58 \\
Ministral-8B-Instruct \cite{jiang2023mistral7b} & 70.30 & 67.11 & 64.71 & 60.02 & 38.91 & 42.49 \\
Qwen3-8B \cite{qwen3technicalreport} & 78.87 & 76.51 & 75.75 & 56.91 & 49.05 & 50.88 \\
\midrule
\multicolumn{7}{c}{\textit{Large Language Models for Reasoning}} \\
DeepSeek-R1-Distill-Llama-8B \cite{guo2025deepseek-r1} & 77.03 & 50.56 & 47.17 & 69.41 & 40.81 & 40.22 \\
Qwen3-8B-think \cite{qwen3technicalreport} & 82.73 & 77.40 & 74.13 & 72.91 & 60.52 & 61.65 \\
DeepSeek-R1-0528-Qwen3-8B \cite{guo2025deepseek-r1} & 83.00 & 77.40 & 77.07 & 64.03 & 66.90 & 58.50 \\
\midrule
\multicolumn{7}{c}{\textit{SOTA Models}} \\
$Qwen3-8B_{SFT}$ & 81.04 & 76.06 & 74.59 & 68.76 & 54.99 & 58.67  \\
Qwen3-Embedding-8B \cite{qwen3embedding} &  91.85 & 91.95 & 91.76 & 73.86 & 69.96 & 
71.64  \\
INSTRUCTOR-CIT \cite{shuang2024INSTRUCTOR-CIT}  &  88.83 & 88.81 & 88.77 & 84.24 & 82.52 & 83.14  \\
\hdashline
\rowcolor{MorandiOrange} ABEX-RAT (Ours) & 93.07 & 92.84 & \textbf{92.82} & 89.99 & 91.88 & \textbf{90.32}  \\
\bottomrule
\end{tabularx}
\end{table*}

\subsection{Stage 3: RAT Classification}
The final stage trains a lightweight multi-layer perceptron (MLP) classifier, $f(\mathbf{x}; \theta)$, using our RAT mechanism. 

\subsubsection{Adversarial Perturbation}
Standard training assumes that test data follows the same distribution as training data. However, in safety reports, minor lexical changes can shift semantics. To improve robustness, we approximate worst-case perturbations $\mathbf{r}_{\text{adv}}$ using the FGM~\cite{fgm}:
\begin{equation}
    \mathbf{r}_{\text{adv}} = \epsilon \frac{\mathbf{g}}{\|\mathbf{g}\|_2}, \quad \text{where } \mathbf{g} = \nabla_{\mathbf{x}} \mathcal{L}(f(\mathbf{x}; \theta), y)
\end{equation}
Here, $\mathbf{g}$ represents the gradient of the loss with respect to the input embeddings. The term $\frac{\mathbf{g}}{\|\mathbf{g}\|_2}$ normalizes the gradient to a unit vector, ensuring the perturbation direction maximizes the loss, while $\epsilon$ strictly bounds the magnitude of this perturbation. This simulates a hard example that lies just outside the model's current decision boundary.

\subsubsection{Stochastic Regularization}
Traditional adversarial training requires calculating gradients for $\mathbf{r}_{\text{adv}}$ at every step, effectively doubling the training time. To mitigate this computational expense, RAT applies perturbations stochastically. For each batch, a Bernoulli trial $k \sim \text{Bernoulli}(p_{\text{rat}})$ determines if the adversarial loss is added:
\begin{equation}
    \mathcal{L}_{\text{total}} = \mathcal{L}_{\text{std}} + k \cdot \mathcal{L}_{\text{adv}}
\end{equation}
where $\mathcal{L}_{\text{std}} = \mathcal{L}(f(\mathbf{x}; \theta), y)$ is the standard loss on clean embeddings, and $\mathcal{L}_{\text{adv}} = \mathcal{L}(f(\mathbf{x} + \mathbf{r}_{\text{adv}}; \theta), y)$ is the adversarial loss on perturbed embeddings. This stochasticity acts as a regularizer, preventing the model from overfitting to the adversarial examples themselves.

\subsubsection{Class-Balanced Loss}
Finally, to further combat the long-tail distribution, we instantiate the loss function $\mathcal{L}$ with the focal loss~\cite{lin2017focal} rather than standard Cross-Entropy. The focal loss reshapes the loss surface to down-weight easy examples and focus on hard negatives:
\begin{equation}
    \text{FL}(p_t) = -\alpha_t (1-p_t)^\gamma \log(p_t)
\end{equation}
where $p_t$ is the model's predicted probability for the ground-truth class. The modulating factor $(1-p_t)^\gamma$ decays to zero as $p_t \to 1$, effectively silencing well-classified examples. By integrating Focal Loss into the RAT objective, our framework robustly handles both input-level perturbations $\mathbf{r}_{\text{adv}}$ and dataset-level imbalance $\alpha_t$ and $\gamma$.

\section{EXPERIMENTS}

\subsection{Experimental Setup}

\noindent\textbf{Dataset and Metrics:} We evaluate the proposed framework on the OSHA dataset\footnote{\url{https://github.com/qiao77/Injury-Narratives}} \cite{qiao2022construction}, a benchmark corpus comprising 4,770 construction accident reports classified into seven distinct categories. To rigorously assess generalization, we employ a stratified random split of 8:1:1 for training, validation, and testing, respectively. This stratification ensures that the label distribution in the test set mirrors the severe real-world imbalance. Accordingly, we report two key metrics: Weighted-F1 to reflect overall classification accuracy dominated by majority classes, and Macro-F1, which treats all classes equally and serves as the primary indicator of performance on long-tail, high-risk incident types.

\noindent\textbf{Implementation Details:} All models are implemented in PyTorch and trained on a single NVIDIA RTX 4090 GPU. The downstream MLP classifier consists of two linear layers with ReLU activation. The hyperparameter configuration is fixed as follows: learning rate $\eta = 1 \times 10^{-4}$, batch size $B=16$, and maximum epochs $E=100$. For the proposed components, the RAT is controlled by a Bernoulli probability $p_{\text{rat}}=0.5$ and a perturbation magnitude $\epsilon=0.1$. The focal loss utilizes a focusing parameter $\gamma=3.0$ to penalize hard-to-classify examples.

\noindent\textbf{Baselines:} To comprehensively benchmark performance, we compare \texttt{ABEX-RAT} against three distinct categories of methods:
\begin{enumerate}
    \item \textit{Traditional Fine-tuning:} Standard discriminative models including static embeddings (\texttt{FastText}) and pre-trained transformers (\texttt{BERT}, \texttt{RoBERTa}) fine-tuned on the raw training data.
    \item \textit{Zero-shot LLMs:} State-of-the-art LLMs (\texttt{Llama3.1-8B-Instruct}, \texttt{Qwen3-8B}) prompted directly for classification without parameter updates, testing their intrinsic domain knowledge.
    \item \textit{Domain-Specific SOTA:} Competitive methods tailored for this task, including the previous state-of-the-art \texttt{INSTRUCTOR-CIT} and a fine-tuned version of Qwen3 (\texttt{Qwen3-8B\textsubscript{SFT}}) with LoRA \cite{hu2022lora}, which represents the brute-force LLM adaptation approach.
\end{enumerate}

\subsection{Main Results and Analysis}

The quantitative comparison results are summarized in Table I. Our proposed \texttt{ABEX-RAT} framework achieves a new state-of-the-art performance on the OSHA dataset, recording a Weighted-F1 of 92.82\% and a Macro-F1 of 90.32\%. These metrics confirm that our synergistic approach robustly handles the dichotomy between frequent and rare accident categories.

\texttt{ABEX-RAT} outperforms the strongest traditional baseline, \texttt{RoBERTa-base}, by 2.0\% in Weighted-F1 and 0.58\% in Macro-F1. While the numerical margin in Macro-F1 appears modest, it is critical to note that our method achieves this using a lightweight MLP on fixed embeddings, avoiding the heavy computational inference cost associated with full Transformer models. The zero-shot performance of general-purpose LLMs (e.g., \texttt{Qwen3-8B}) is substantially lower, lagging behind supervised methods by over 40\% in Macro-F1. This underscores that despite their linguistic capabilities, generalist LLMs lack the specific taxonomic knowledge required for precise occupational safety classification, validating the need for task-specific supervision.

A pivotal finding is that \texttt{ABEX-RAT} surpasses the fine-tuned \texttt{Qwen3-8B\textsubscript{SFT}} by a remarkable margin. This result suggests that on small, imbalanced datasets, brute-force fine-tuning of billions of parameters leads to severe overfitting on majority classes. In contrast, our strategy of freezing the LLM backbone + augmenting the data space + smoothing the decision boundary proves to be a far more effective and data-efficient paradigm. Furthermore, the 18.68\% jump in Macro-F1 over the \texttt{Qwen3-Embedding} baseline explicitly isolates the contribution of our contributions, proving that the high performance stems from the proposed augmentation and adversarial training rather than merely the quality of the base embeddings.

\begin{figure}
\centering
\includegraphics[width=0.45\textwidth]{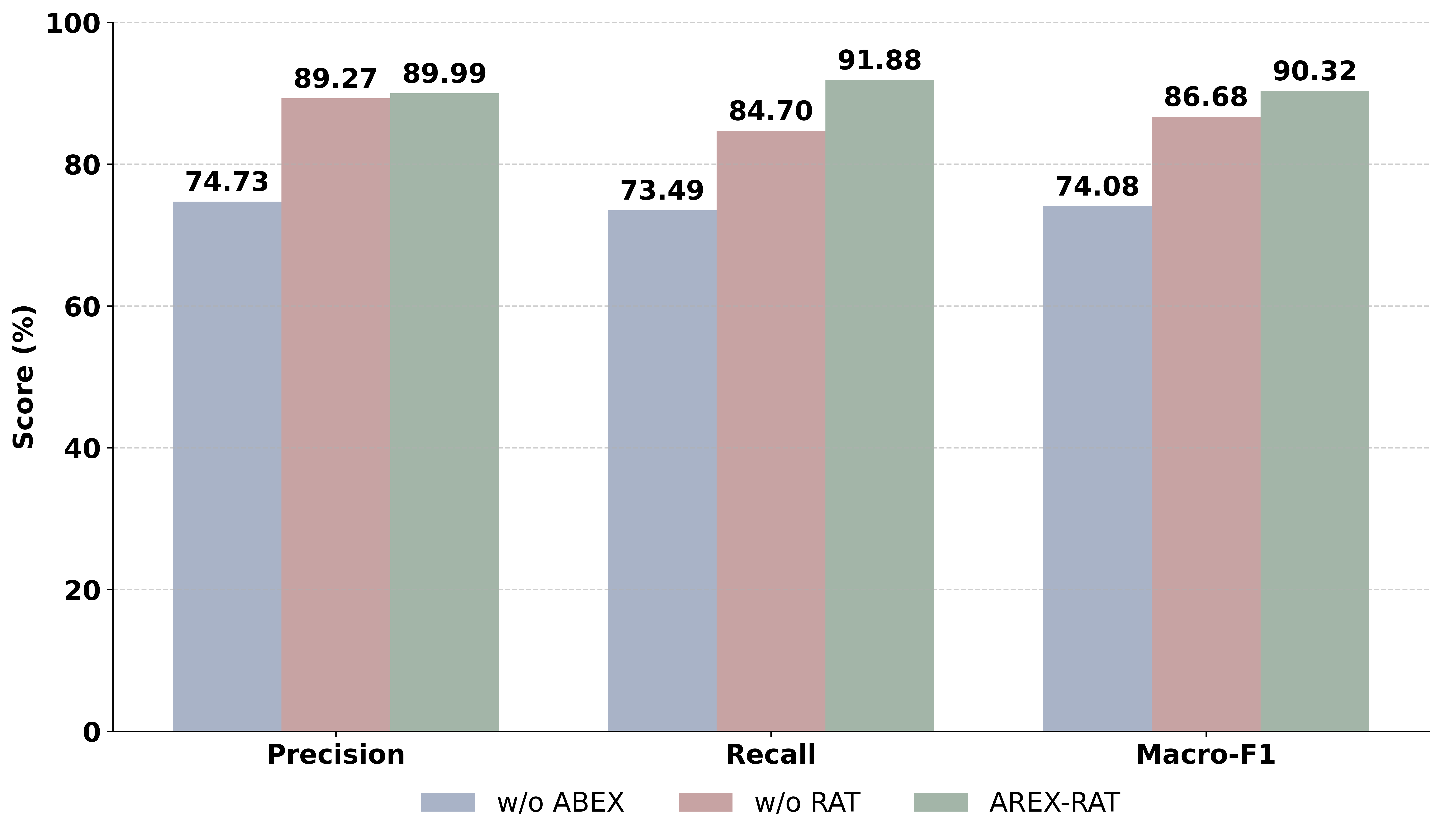}
\caption{Results of the ablation experiment.}
\end{figure}

\section{DISCUSSION}

\subsection{Ablation Study: Decoupling Synergy}

To disentangle the contributions of the ABEX data augmentation and the RAT protocol, we conducted a rigorous ablation study. We compared our full \texttt{ABEX-RAT} model against two variants: (1) \texttt{w/o RAT}, which utilizes the augmented data but trains with standard cross-entropy loss; and (2) \texttt{w/o ABEX}, which applies RAT on the original, imbalanced dataset. 

The results, visualized in Figure 4, reveal a clear synergistic effect. The \texttt{w/o RAT} model achieves a strong macro-F1 score of 86.68\%, demonstrating that ABEX successfully injects necessary semantic diversity into minority classes. However, without RAT, the model tends to overfit to specific linguistic patterns in the synthetic data. In contrast, the \texttt{w/o ABEX} model yields a significantly lower macro-F1 of 74.08\%. This confirms that while adversarial training enhances robustness, it cannot invent missing knowledge—it struggles when the underlying data manifold is sparsely populated.

Our full framework outperforms both, achieving the highest macro-F1 of 90.32\%. This synergy is most evident in the Recall metric, which jumps from 84.70\% to 91.88\%. This suggests that RAT effectively smooths the decision boundary around the synthetic samples provided by ABEX, allowing the model to generalize better to unseen, challenging minority-class instances.

\subsection{Fine-grained Performance Analysis}

\begin{figure}
\centering
\includegraphics[width=0.45\textwidth]{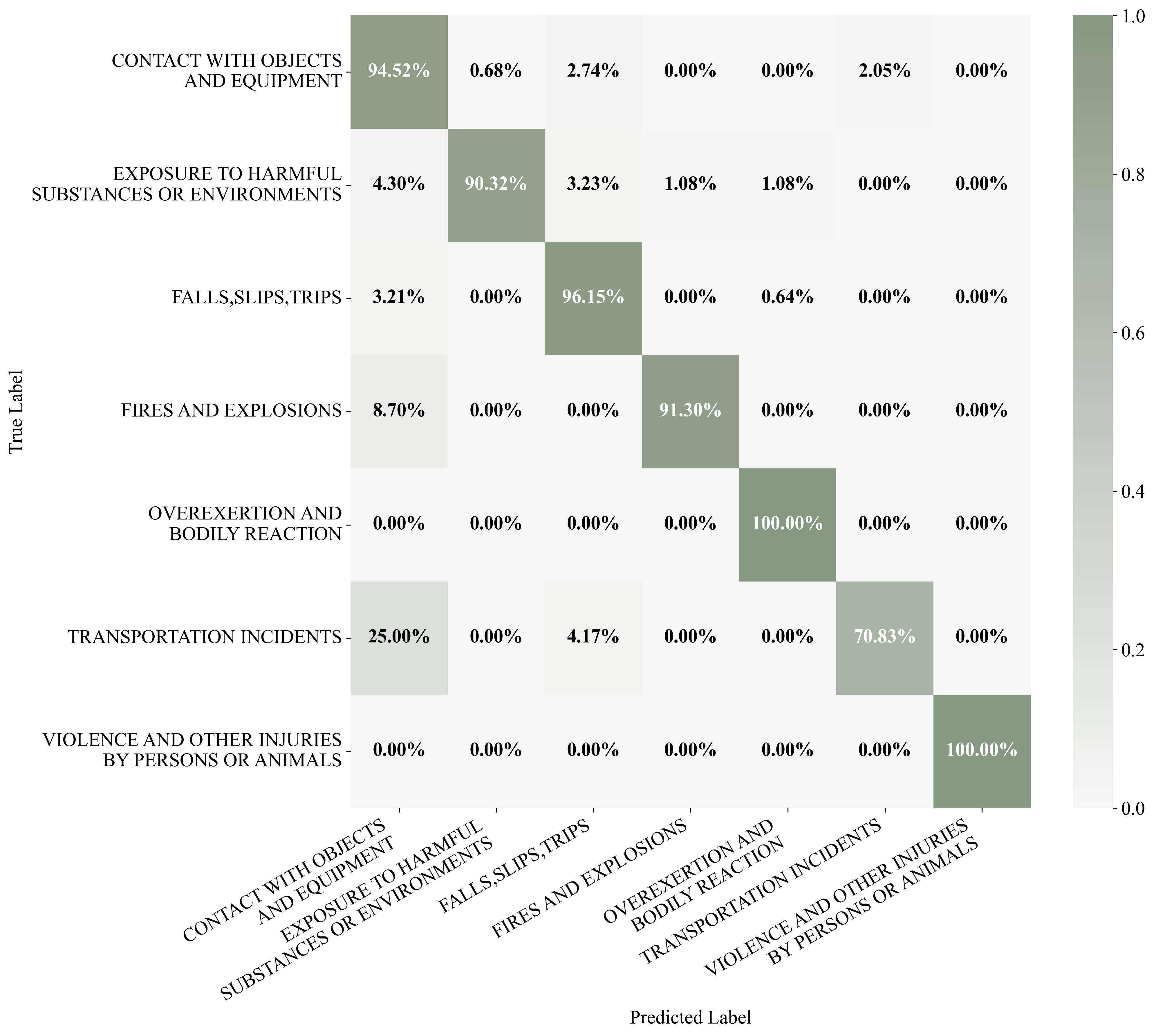}
\caption{Normalized confusion matrix.}
\end{figure}

To further dissect classification performance, we analyze the normalized confusion matrix (Figure 5). The diagonal elements demonstrate robust recall across all categories, specifically distinguishing the framework's efficacy on the rarest class, \textit{FIRES AND EXPLOSIONS} (91.30\%). This is a critical improvement over baseline methods which often ignore this category entirely due to sample scarcity. The analysis also reveals semantically plausible misclassifications. For instance, 25\% of \textit{TRANSPORTATION INCIDENTS} are predicted as \textit{CONTACT WITH OBJECTS}. A manual inspection reveals these often involve vehicles striking pedestrians, which sits at the semantic intersection of both categories. This suggests that future work could benefit from multi-label modeling or hierarchical classification to resolve such ambiguities.

\subsection{Case Study}

To intuitively understand the improvements, we visualize real-world examples in Figure \ref{fig: cases}.

\textbf{Case 1:} The upper panel demonstrates the ABEX pipeline. The abstractive step successfully strips dates, names while retaining the causal chain. The expansion step then generates a grammatically distinct variation that preserves these semantics, effectively enriching the feature space for the \textit{Fire} category.

\textbf{Case 2:} The lower panel illustrates a hard sample involving a forklift. The baseline \texttt{RoBERTa} model, likely biased by the high-frequency keyword forklift, incorrectly predicts \textit{TRANSPORTATION INCIDENT}. However, \texttt{ABEX-RAT} correctly identifies the root cause as \textit{FALLS, SLIPS, TRIPS}. This indicates that our model has learned to attend to the mechanism of injury rather than just surface-level nouns, validating the robustness gain from adversarial training.

\begin{figure}[h]
    \centering
    \begin{outerbox}[Case Study Analysis]
        
        \begin{promptbox}[mBlue]{Case 1: ABEX Generative Process}
            \textbf{[Original Text]} On August 12, Employee \#1 was cutting a pipe... sparks from the torch ignited residual gas...
            
            \tcbline
            
            \textbf{[Abstraction]} An employee was using a torch to cut piping. Sparks ignited flammable residue, causing a flash fire.
            
            \tcbline
            
            \textbf{[Augmented Sample]} \textit{While performing hot work on a pipeline, a worker triggered a flash fire when torch sparks contacted combustible fumes.}
        \end{promptbox}

        \begin{promptbox}[mRed]{Case 2: Error Correction }
            \textbf{[Input Text]} Employee \#1 was standing on the tines of a \underline{forklift} to reach a shelf. He lost balance and \underline{fell approx 7 feet} to the concrete.
            
            \tcbline
            
            \begin{minipage}[t]{0.48\linewidth}
                \textcolor{mRed}{\textbf{$\times$ Baseline (RoBERTa):}}\\
                \textit{Transportation Incidents}\\
                \scriptsize{(Error: Over-reliance on "forklift")}
            \end{minipage}
            \hfill
            \begin{minipage}[t]{0.48\linewidth}
                \textcolor{mBlue}{\textbf{$\checkmark$ Ours (ABEX-RAT):}}\\
                \textit{Falls, Slips, Trips}\\
                \scriptsize{(Correct: Captures "fell 7 feet")}
            \end{minipage}
        \end{promptbox}
        
    \end{outerbox}

    \caption{Case Study of the proposed framework. \textbf{Top (Blue):} Demonstrates the semantic consistency of the ABEX generation pipeline. \textbf{Bottom (Red):} Highlights the model's robustness in correcting hard samples where baselines fail due to keyword bias.}
    \label{fig: cases}
\end{figure}

\section{Conclusion}
In this paper, we presented ABEX-RAT, a data-efficient framework tailored for the high-stakes task of occupational accident report classification. By synergizing an LLM-driven abstractive-expansive data augmentation pipeline with a lightweight RAT protocol, we effectively mitigated the dual challenges of extreme data scarcity and class imbalance. Empirical evaluations on the OSHA dataset confirm that ABEX-RAT establishes a new state-of-the-art, achieving a Macro-F1 score of 90.32\% and significantly outperforming both traditional baselines and computationally expensive large model fine-tuning. These findings validate that targeted data enrichment combined with robust regularization offers a superior, resource-efficient alternative to brute-force model scaling for specialized domain applications. Future research will extend this paradigm to multi-label scenarios and explore hierarchical classification techniques to better resolve semantic overlaps between closely related accident categories.

\section*{Acknowledgment}

The authors would like to thank the anonymous reviewers for their helpful comments and suggestions.

\bibliographystyle{unsrt}
\bibliography{ref}

\end{document}